\documentclass[10pt,journal,compsoc]{IEEEtran}

\usepackage{booktabs}    
\usepackage{arydshln}    
\usepackage{colortbl}   
\usepackage{xcolor}     
\usepackage{array}      

\usepackage{cite}

\ifCLASSINFOpdf
  \usepackage[pdftex]{graphicx}
  \DeclareGraphicsExtensions{.pdf,.jpeg,.png}
\else
  \usepackage[dvips]{graphicx}
\fi

\usepackage{amsmath}
\usepackage{amssymb}

\usepackage{array}

\usepackage[caption=true,font=footnotesize]{subfig}

\usepackage{stfloats}

\ifCLASSOPTIONcaptionsoff
  \usepackage[nomarkers]{endfloat}
 \let\MYoriglatexcaption\caption
 \renewcommand{\caption}[2][\relax]{\MYoriglatexcaption[#2]{#2}}
\fi

\begin{document}

\title{Improving Cardiac Risk Prediction Using Data Generation Techniques}

\author{Alexandre~Cabodevila,
        Pedro~Gamallo-Fernández,
        Juan~C.~Vidal,
        and~Manuel~Lama
\thanks{
Alexandre Cabodevila, Pedro Gamallo-Fernández, Juan C. Vidal, and Manuel Lama 
are with the Centro Singular de Investigación en Tecnoloxías Intelixentes (CiTIUS),
Universidade de Santiago de Compostela, Santiago de Compostela, SPAIN
e-mail: \{alexandre.cabodevila.lopez, pedro.gamallo.fernandez, juan.vidal, manuel.lama\}@usc.es }
}

\IEEEtitleabstractindextext{%
\begin{abstract}
Cardiac rehabilitation constitutes a structured clinical process involving multiple interdependent phases, individualized medical decisions, and the coordinated participation of diverse healthcare professionals. This sequential and adaptive nature enables the program to be modeled as a business process, thereby facilitating its analysis. Nevertheless, studies in this context face significant limitations inherent to real-world medical databases: data are often scarce due to both economic costs and the time required for collection; many existing records are not suitable for specific analytical purposes; and, finally, there is a high prevalence of missing values, as not all patients undergo the same diagnostic tests. To address these limitations, this work proposes an architecture based on a Conditional Variational Autoencoder (CVAE) for the synthesis of realistic clinical records that are coherent with real-world observations. The primary objective is to increase the size and diversity of the available datasets in order to enhance the performance of cardiac risk prediction models and to reduce the need for potentially hazardous diagnostic procedures, such as exercise stress testing. The results demonstrate that the proposed architecture is capable of generating coherent and realistic synthetic data, whose use improves the accuracy of the various classifiers employed for cardiac risk detection, outperforming state-of-the-art deep learning approaches for synthetic data generation.
\end{abstract}

}

\maketitle

%
\IEEEpeerreviewmaketitle

\section{Introduction}
%
%
%
%
Cardiovascular diseases (CVDs), or heart diseases, constitute the \textbf{leading cause of death} worldwide, according to data provided by the World Health Organization (WHO), accounting for approximately 18 million deaths each year~\cite{who2024cvd}. This trend is also observed in Europe, where these conditions are responsible for nearly 10,000 deaths per day, making them the primary cause of premature mortality across the continent~\cite{who2024europe}. In addition, these disorders, including myocardial infarction and cerebrovascular accidents, represent a substantial burden on public healthcare systems and on the global economy. Consequently, they constitute one of the main areas of research in the medical field, fostering continuous advances in their prevention and treatment.

One of the major challenges in combating these pathologies lies in their early detection, as symptoms often appear only once significant damage or deterioration has already occurred. 
As a result, the chances of implementing effective treatment are considerably reduced. Owing to this need for timely diagnosis, cardiac risk prediction has emerged as a fundamental tool in preventive medicine: if at-risk patients can be identified early, it becomes possible to intervene in order to prevent further complications, improve patients’ quality of life, and reduce mortality rates associated with heart disease.

Despite efforts aimed at early risk detection, this task remains challenging, as it requires frequent assessments and regular visits to healthcare centers or hospitals. This contrasts with real-world behavior, where most patients do not seek medical attention until symptoms become apparent~\cite{sierradelsierra2008}. Consequently, one of the most widely adopted strategies to address cardiovascular diseases is the implementation of \textbf{cardiac rehabilitation programs}, targeted at patients who have experienced a cardiac event or who present risk factors such as hypertension or diabetes. These programs are highly valued by the medical community, as they have repeatedly demonstrated their effectiveness in reducing the risk of future cardiovascular events and mortality, while also improving participants’ quality of life~\cite{Anderson2016,Taylor2004,Rauch2016}.

Such programs typically consist of a series of in-person sessions held at a fixed frequency, usually once or twice per week, during which patients engage in physical exercise following individualized guidelines and under the supervision of a cardiologist. The specialist is responsible for adjusting training intensity according to the patient’s clinical condition. Furthermore, scientific organizations such as the American Heart Association emphasize the importance of incorporating strategies aimed at addressing risk factors such as hypercholesterolemia and hypertension~\cite{AHA2024}. Another fundamental component is psychological care and support, which helps patients adapt to their new circumstances and improves adherence to the rehabilitation program~\cite{Hughes2022Psychosocial}.

Based on the large volume of data collected throughout these programs, the problem of cardiac risk prediction can be addressed at different stages, although it is most commonly performed at the end of the rehabilitation process. In this way, healthcare professionals can assess whether the protocol has been effective in improving the patient’s condition or whether additional measures are required to reverse the situation. Moreover, such prediction could eliminate the need for an exercise stress test (ergometry), which is particularly advantageous given that these procedures are associated with certain risks, especially in high-risk patients, and may potentially lead to severe cardiac events.

\subsection{Modeling the program as a business process}
Business processes, although often unnoticed, are present in a wide range of domains, such as banking institutions or the management of online retail platforms. Process mining is a discipline whose objective is to extract information related to processes from event data generated during their execution, which are stored in so-called event logs. These logs are datasets that contain information about the historical executions of a given process. The knowledge extracted from them can be used to discover a process model, improve existing processes, or verify whether they are being executed as intended~\cite{DBLP:books/sp/Aalst16}.

An \textbf{event log} exclusively contains data related to the different executions of a single process. Each event recorded in the log is ordered and associated with the execution of a specific activity, such as \textit{incident registration} or \textit{incident analysis}. These activities correspond to the individual actions or \textbf{tasks} that compose the process. Moreover, each event in the log must be linked to a unique execution instance of the process, known as a \textbf{case} or trace. A case represents an ordered and complete execution of all the activities or steps that constitute a business process.

A cardiac rehabilitation program can be understood and modeled as a business process, since it consists of a set of structured and sequential activities carried out to achieve a common goal: the recovery and improvement of the patient’s health. These activities correspond to clinical consultations in which the participant’s condition is monitored, hospital admissions, and the occurrence of cardiac diseases or physiological alterations (i.e., changes in clinical parameters). In this setting, the evolution of each patient constitutes a trace, associated with both global variables, those that remain constant across events, such as family history, sex, or height, and event-level attributes, such as the clinical measurements obtained by healthcare professionals. This representation makes it possible to design a structure that captures the clinical process in a sequential manner, with the ultimate aim of achieving more accurate cardiac risk prediction.

\subsection{Cardiac risk prediction techniques}
Early approaches to cardiac risk prediction were predominantly based on \textbf{multivariate statistical models}, developed from large population cohorts and relying on a limited set of classical risk factors, such as age, blood pressure, cholesterol levels, and smoking habits. Among the most influential examples is the \textbf{Framingham Risk Score} (FRS)~\cite{Wilson1998Prediction}, derived from the Framingham Heart Study and originally conceived to estimate the 10-year risk of coronary heart disease, later extended to encompass a broader range of cardiovascular outcomes~\cite{dagostino2008general}. Despite its widespread adoption, concerns have been raised regarding its generalizability beyond the population in which it was developed.

Subsequent models sought to address these limitations by incorporating regional and population-specific information. The \textbf{SCORE} model~\cite{conroy2003score}, for instance, estimates 10-year cardiovascular mortality risk based on European cohorts and provides risk charts adapted to different epidemiological contexts, although it does not account for non-fatal events. Similarly, the \textbf{QRISK} and \textbf{QRISK2} models~\cite{qrisk,qrisk2}, developed using clinical data from the United Kingdom, extended earlier formulations by including additional factors such as body mass index, ethnicity, family history, and socioeconomic status, thereby improving risk estimation for both fatal and non-fatal cardiovascular events. Other notable contributions include the \textbf{Reynolds Risk Score}~\cite{ridker2007reynolds} and the \textbf{PROCAM}~\cite{assmann2002procam}, which further enriched traditional approaches through the integration of novel clinical variables.

While these statistical models have played a pivotal role in identifying major cardiovascular risk factors and remain valued for their interpretability, they are inherently limited by their reliance on a small number of variables and their assumption of linear and static relationships. Such constraints hinder their ability to capture the complex, multifactorial, and dynamic nature of cardiovascular disease, reducing their adaptability to heterogeneous populations and evolving clinical settings.

In response to these challenges, \textbf{semi-parametric statistical models} have been proposed as a compromise between interpretability and flexibility. By relaxing restrictive assumptions, such as linearity or fully specified baseline hazard functions, these methods enable more realistic modeling of cardiovascular risk while retaining a structure amenable to clinical interpretation.

Representative examples include the Cox proportional hazards model, which has been applied in a semi-parametric manner to estimate cardiovascular risk without explicitly defining the baseline hazard~\cite{Moons2012}, as well as extensions incorporating penalized or fractional splines to model non-linear and time-varying effects of key risk factors~\cite{Abrahamowicz1997}. Generalized additive models (GAMs) further extend this paradigm by capturing complex non-linear relationships through smooth functions, as demonstrated in recent studies addressing environmental and cardiovascular interactions~\cite{Zhu2024}.

In summary, semi-parametric approaches mitigate several limitations of classical statistical models by accommodating non-linear effects, integrating multiple risk factors, and preserving clinical interpretability. Nevertheless, the increasing availability of large-scale and high-dimensional clinical data, together with the complexity of underlying biological interactions, has revealed limitations in terms of predictive performance and scalability. Within this context, \textbf{machine learning and deep learning–based models} have emerged as promising alternatives, offering enhanced capacity to capture complex patterns and to automate feature selection in modern cardiovascular risk prediction tasks.
\subsection{Need for synthetic data generation}
Despite the advantages offered by these models, their practical application faces a major limitation: a \textbf{strong dependence on large volumes of high-quality data}. When such data are scarce or incomplete, the amount of useful knowledge that can be extracted is substantially reduced. This situation is common in real-world clinical settings, where the number of patients relevant to a given study is inherently limited. Furthermore, in practice, not all variables are available for every individual, as patients do not undergo the same tests or examinations. As a result, the amount of data that can be used to train predictive models is considerably smaller than it might initially appear. This scarcity is further exacerbated by the slow pace at which clinical data are collected. Considering a cardiac rehabilitation program with weekly sessions and 200 patients participating simultaneously, it would take an entire month to generate 800 data records, assuming that all participants attend every session and that all relevant measurements are collected each time. Even this assumption is optimistic, as not all patients with cardiovascular disease will enroll in the program and, among those who do, not all will consent to data sharing. Consequently, the final number of available records per unit of time is relatively low.

In light of these constraints, the need arises to employ \textbf{synthetic data generation techniques} in order to expand the size of the available dataset and enhance the predictive performance and robustness of artificial intelligence–based models. These techniques aim to simulate new, realistic clinical examples that preserve the statistical relationships present in the original data, without directly replicating them.

The application of data generation techniques to increase sample size and improve model robustness poses several fundamental challenges:
\begin{itemize}
  \item \textbf{Data quality:} Synthetic data must exhibit a level of quality comparable to that of the original data, as poorly generated samples may degrade model performance.
  \item \textbf{Noise and overfitting:} If synthetic data are excessively similar to the original samples, they may induce overfitting; if they are too dissimilar, they may introduce noise and hinder learning.
  \item \textbf{Distribution preservation:} Generated data should respect the original data distribution in order to avoid introducing biases that lead models to learn unrealistic patterns.
  \item \textbf{Semantic correctness:} It is essential that synthetic data satisfy basic constraints and remain clinically coherent. For instance, a patient who has suffered a myocardial infarction should exhibit elevated values in the corresponding cardiac biomarkers~\cite{DBLP:conf/mlhc/ChoiBMDSS17}.
\end{itemize}

This work proposes a novel approach to synthetic data generation for cardiac rehabilitation programs, aimed at improving cardiac risk prediction at program completion. Accurate risk estimation at this stage may reduce the reliance on exercise stress testing, a commonly used but potentially risky procedure, thereby contributing to safer and more efficient patient assessment based on routinely collected clinical data.

The main contribution is the design and implementation of a \textbf{Conditional Variational Autoencoder} (CVAE) architecture tailored to clinical rehabilitation data. Its conditional formulation enables controlled generation of synthetic patient records consistent with predefined clinical characteristics, such as risk status, ensuring semantic coherence and suitability for downstream predictive tasks.

In addition, the proposed method incorporates contrastive learning, sparsity-inducing regularization, and a guided latent-space interpolation strategy to improve latent space structure, sample diversity, and robustness in low-data and class-imbalanced settings.

The approach is evaluated through an indirect validation framework, measuring its impact on multiple cardiac risk classifiers trained with augmented datasets. Comparisons with state-of-the-art generative models~\cite{DBLP:conf/nips/XuSCV19, DBLP:journals/ai/WangN25, DBLP:conf/nips/GulrajaniAADC17} show consistent improvements in predictive performance, demonstrating the effectiveness of the proposed data generation strategy.

\section{State of the Art}
Synthetic data generation is widely adopted in machine learning settings where real data are scarce or highly imbalanced. In such scenarios, \textbf{data augmentation} is employed to artificially increase dataset size and diversity by generating new samples that preserve the coherence and semantic validity of the original data, thereby improving model generalization and mitigating overfitting.

Classical augmentation techniques include the \textbf{addition of random noise} to numerical variables~\cite{DBLP:conf/ssci/Moreno-BareaSJU18,DBLP:journals/jbd/ShortenK19} and \textbf{interpolation between samples}, where new instances are generated as intermediate points in the feature space~\cite{DBLP:journals/jair/ChawlaBHK02, DBLP:conf/ijcnn/HeBGL08}. Although initially developed for image-based tasks~\cite{DBLP:journals/jbd/ShortenK19, DBLP:journals/corr/abs-1712-04621}, these methods have also been adapted to tabular data. However, such approaches present notable limitations: they do not explicitly model the underlying data distribution, lack contextual awareness, are sensitive to outliers, and often generate unrealistic or overly similar samples, resulting in limited effective diversity~\cite{YANG2024110204}.

To overcome these drawbacks, neural network–based generative models have gained prominence, including \textbf{VAEs}~\cite{DBLP:journals/corr/KingmaW13}, \textbf{GANs}~\cite{DBLP:conf/nips/GoodfellowPMXWOCB14}, \textbf{Diffusion Models}~\cite{DBLP:conf/icml/Sohl-DicksteinW15}, and \textbf{Generative Transformers}~\cite{DBLP:conf/nips/VaswaniSPUJGKP17}. In this work, the focus is placed on VAEs and GANs, as they are the most widely used across multiple data modalities, including tabular and sequential data. While diffusion models have recently shown strong performance for tabular data, notably through TabDDPM~\cite{DBLP:conf/icml/KotelnikovBRB23}, their high computational cost and architectural complexity limit their practical applicability compared to VAEs and GANs.

GAN-based models have demonstrated remarkable success in generating highly realistic samples across domains such as images~\cite{DBLP:conf/cvpr/LedigTHCCAATTWS17}, audio~\cite{DBLP:conf/iclr/DonahueMP19}, and video~\cite{DBLP:conf/nips/GulrajaniAADC17}, with adaptations also proposed for tabular and sequential data~\cite{DBLP:journals/corr/abs-1811-11264, DBLP:conf/nips/GulrajaniAADC17}. Their adversarial training scheme enables high-fidelity sample generation but often suffers from training instability, mode collapse, and the absence of an explicit probabilistic formulation, which complicates controlled data generation.

Variational Autoencoders constitute a widely adopted alternative, with applications spanning image synthesis~\cite{DBLP:conf/nips/KingmaMRW14}, time series~\cite{DBLP:journals/corr/abs-2111-08095}, and tabular data~\cite{DBLP:journals/ai/WangN25}. VAEs learn a probabilistic latent space that enables coherent sample generation through random sampling and decoding. The continuity of this latent space facilitates interpolation and exploration of unobserved regions while maintaining semantic validity. Compared to GANs, VAEs typically exhibit more stable training behavior, although they may struggle to capture highly complex data distributions and and do not explicitly enforce the preservation of specific data characteristics.

\textbf{Conditional generation} addresses these limitations by enabling explicit control over the characteristics of the generated samples, which is particularly advantageous in imbalanced datasets and medical applications. Conditional variants such as \textbf{CVAEs}~\cite{DBLP:conf/nips/SohnLY15} and \textbf{CGANs}~\cite{DBLP:journals/corr/MirzaO14} incorporate auxiliary information, such as class labels, into the generation process, ensuring semantic consistency between samples and their associated conditions. As a result, CVAE-based~\cite{DBLP:conf/emnlp/GaoBLLZS19, DBLP:journals/kais/WangLDVN25} and CGAN-based~\cite{DBLP:conf/nips/XuSCV19} architectures are commonly adopted when controlled synthetic data generation is required.

In the medical domain, generative models have been predominantly applied to imaging data, including magnetic resonance imaging (MRI), computed tomography (CT) and radiographs, for tasks such as data augmentation, noise reduction, and pathology simulation~\cite{DBLP:journals/cmig/ArmaniousJFKHNG20, DBLP:journals/access/HanRANFMNH19, DBLP:journals/corr/abs-1806-07777}. In contrast, approaches targeting tabular or sequential clinical data remain relatively limited, with examples such as MedDiff~\cite{DBLP:journals/corr/abs-2302-04355} and CEHR-GPT~\cite{DBLP:journals/corr/abs-2402-04400}. These methods, however, do not explicitly model clinical processes as structured sequences of events.

Overall, this analysis highlights that deep learning–based generative models consistently outperform classical techniques, particularly when conditional generation is supported. At the same time, ensuring that generated samples remain coherent with real-world clinical observations remains a central challenge, guiding the selection of the generative architecture adopted in this work.

\section{Theoretical Framework}
\begin{figure*}[ht]
  \centering
  \subfloat[Autoencoder]{%
    \includegraphics[width=0.42\linewidth]{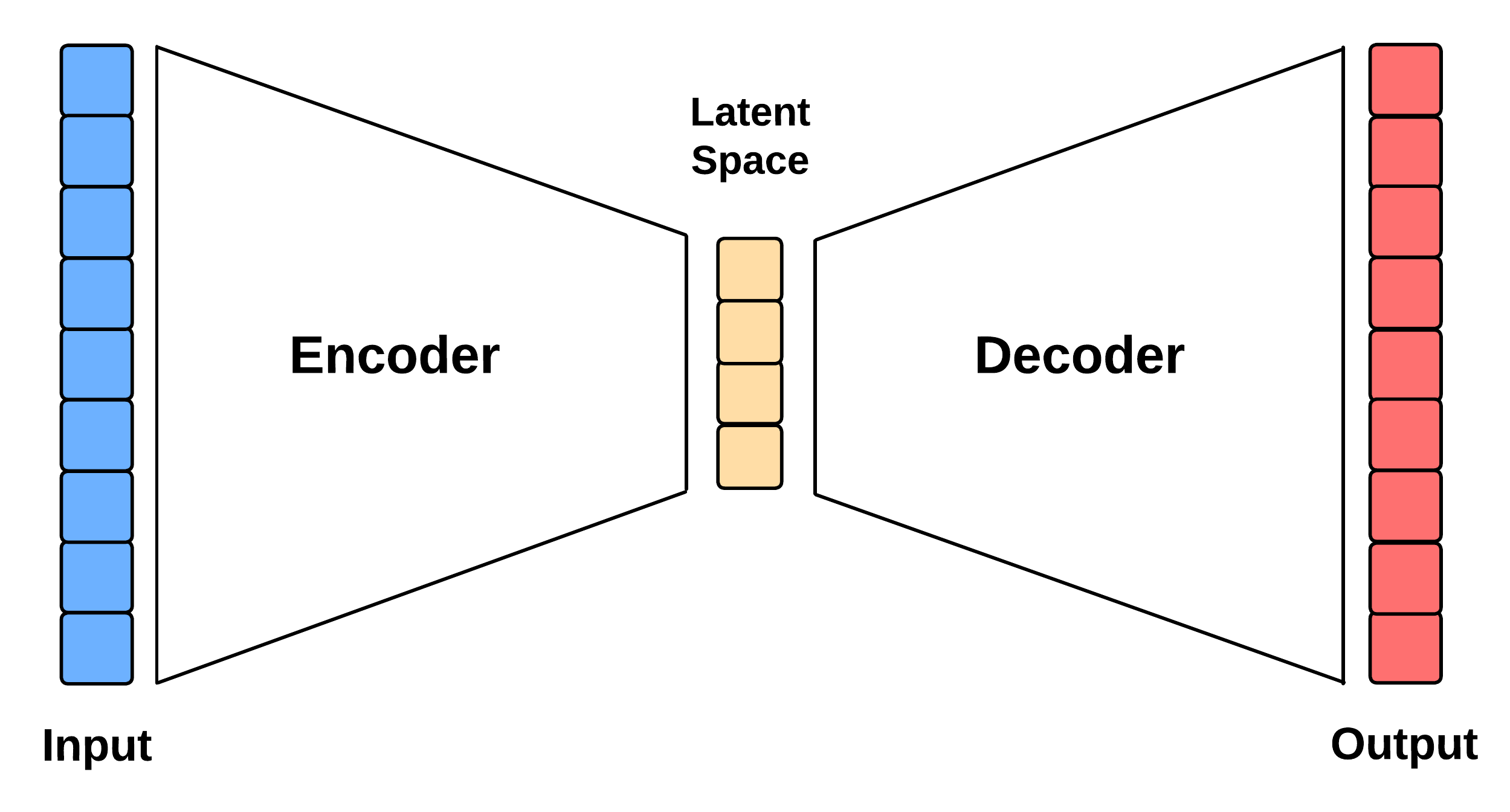}%
    \label{fig:ae}%
  }
  \hfill
  \subfloat[Variational autoencoder]{%
    \includegraphics[width=0.50\linewidth]{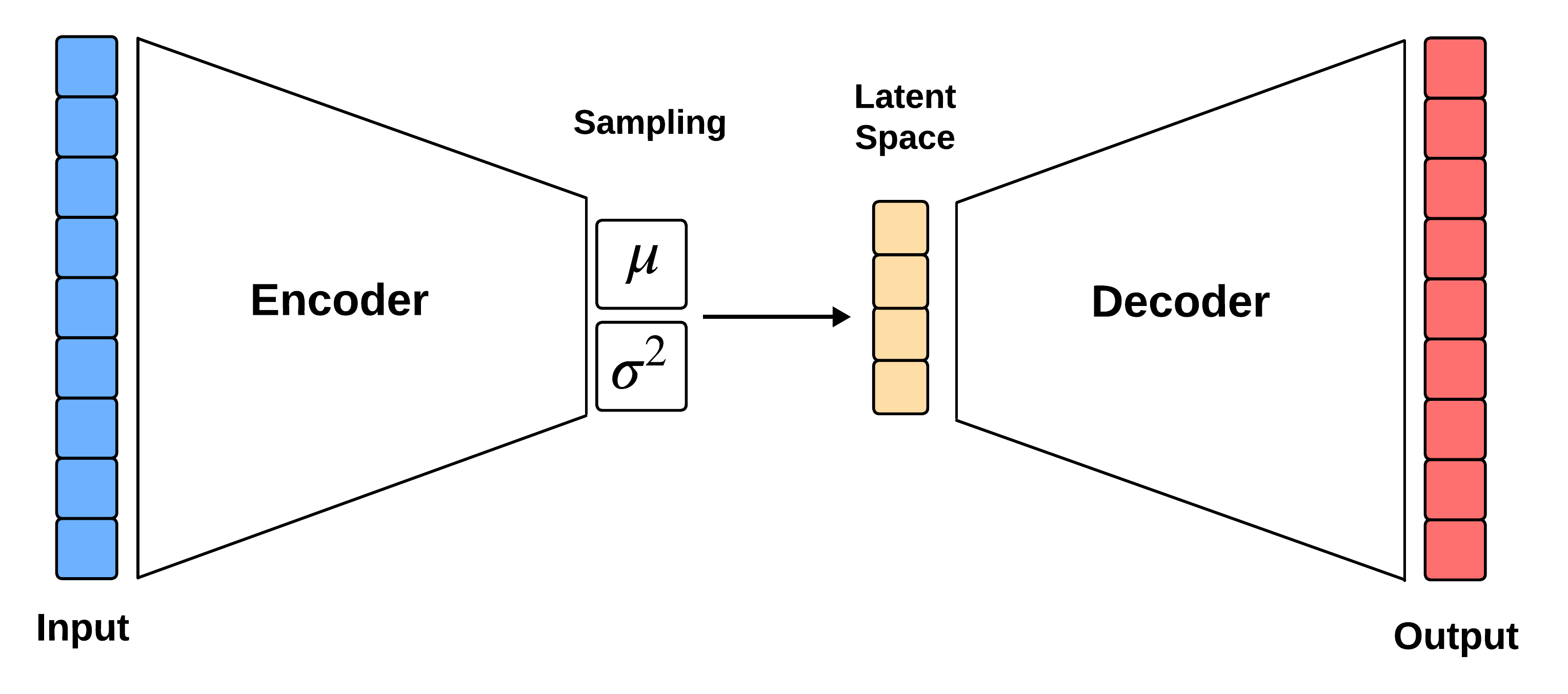}%
    \label{fig:vae}%
  }
  \caption{Comparative schematics of AE and VAE.}
  \label{fig:comparison}
\end{figure*}

\textbf{Autoencoders (AEs)} are a class of feedforward neural networks trained on unlabeled data, that is, in an unsupervised manner~\cite{DBLP:journals/ai/Hinton89}. The AE architecture consists of two main components, each of which constitutes a complete neural network in its own right: an encoder and a decoder. The role of the encoder is to transform the input data into a reduced-dimensional representation, known as the latent space, where the most relevant characteristics of the data are captured. The decoder, in turn, receives this latent representation and attempts to reconstruct the original input from it. Figure~\ref{fig:ae} illustrates the basic structure of an AE.

During training, an AE seeks to minimize the difference between the input data and their reconstructed version, using a loss function that measures the discrepancy between the two. In this way, the AE learns to represent the input efficiently within the latent space, retaining only the most informative aspects of the data.

However, AEs exhibit important limitations with respect to generative capability, as their latent space is not explicitly structured, making the generation of new realistic samples challenging. The organization of this space depends on the data, the dimensionality of the latent representation and the encoder architecture; therefore, it is not always suitable for generation. Moreover, AEs may suffer from overfitting, memorizing characteristics of the training data rather than capturing generalizable patterns.

To overcome the limitations of classical AEs, \textbf{Variational Autoencoders (VAEs)} were introduced~\cite{DBLP:journals/corr/KingmaW13} (Figure~\ref{fig:vae}). Instead of producing a single latent representation, VAEs learn the parameters of a Gaussian distribution, namely, the mean and the logarithm of the variance, from which a latent sample is drawn through a stochastic sampling process. VAEs thus learn a regularized latent distribution, which improves generalization and mitigates overfitting. This probabilistic latent space endows VAEs with two key properties for data generation: \textbf{continuity} (nearby points in the latent space produce similar reconstructions) and \textbf{completeness} (any point sampled from the distribution yields a plausible data instance).

The objective of VAE training is to maximize the Evidence Lower Bound (ELBO), defined in Equation~\ref{eq:elbo}, which balances reconstruction accuracy with latent space regularization in order to ensure a continuous and generative latent representation.
\begin{equation}
\begin{split}
\log p(x) \geq \\
\mathbb{E}_{q_{\phi}(z|x)} &[\log p_{\theta}(x|z)] - \beta \, \text{KLD}(q_{\phi}(z|x) \parallel p(z))
\end{split}
\label{eq:elbo}
\end{equation}

The first term of the ELBO corresponds to the reconstruction error, which encourages the decoder \( p_{\theta}(x|z) \) to accurately reproduce the input \( x \) from the latent vector \( z \). The second term is a regularization penalty that measures the divergence between the approximate posterior distribution learned by the encoder, \( q_{\phi}(z|x) \), and a prior distribution \( p(z) \), typically a standard normal distribution. This deviation is quantified using the Kullback--Leibler divergence (KLD)~\cite{kullback1951information}. The parameter \( \beta \) controls the relative weight of the KLD term. Larger values of \( \beta \) enforce a latent space that more closely matches the prior distribution, thereby improving generalization at the expense of reconstruction fidelity, whereas smaller values favor more accurate reconstructions while placing less emphasis on latent structure.

\begin{figure}[tb]
  \centering
  \includegraphics[width=0.47\textwidth]{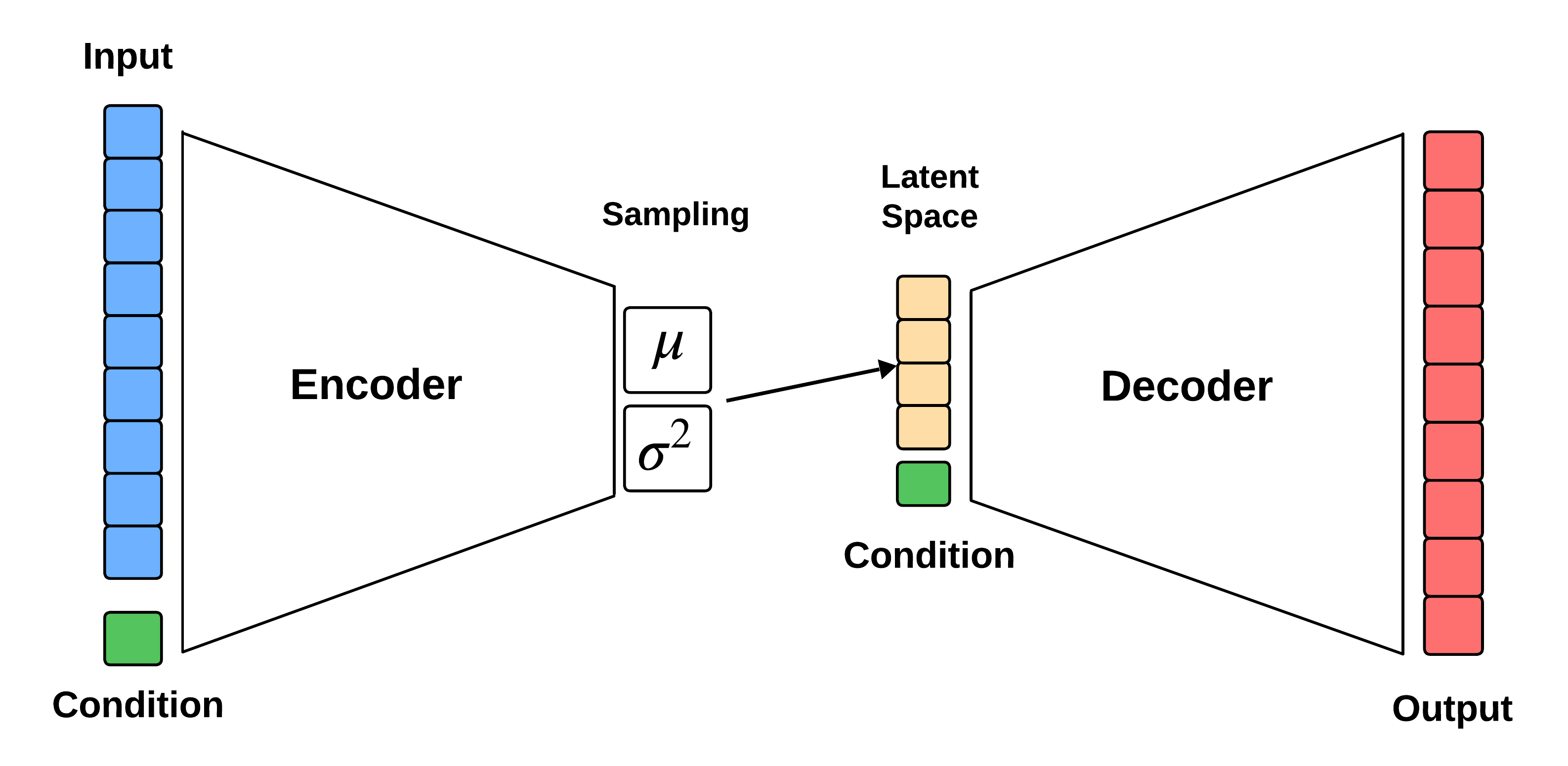}
  \caption{Conditional Variational Autoencoder.}
  \label{fig:cvae}
\end{figure}

\textbf{Conditional Variational Autoencoders (CVAEs)}~\cite{DBLP:conf/nips/SohnLY15}, which form the basis of the main proposal, represent an extension of VAEs that enables explicit control over the type of generated samples (conditional generation), addressing one of the key limitations of standard VAEs. While a VAE can generate arbitrary samples from the data domain (e.g., any image of a handwritten digit), a CVAE allows the specification of a desired output, such as a ``3'' or a ``7'', by incorporating a conditioning variable that guides both encoding and generation. This condition, typically a label or a specific attribute, steers the semantic content of the generated sample, while the latent space is reserved for more subtle variations. Figure~\ref{fig:cvae} provides an overview of the general architecture of a CVAE.

The operation of a CVAE closely resembles that of a VAE, with the key distinction that the conditioning variable \( y \) is incorporated into both the encoder and the decoder. Given an input sample \( x \) and a condition \( y \), the encoder \( q_{\phi}(z|x,y) \) projects the data into a condition-dependent latent space, while the decoder \( p_{\theta}(x|z,y) \) uses both the latent vector \( z \) and the condition \( y \) to reconstruct the input. The prior distribution over the latent space, \( p(z|y) \), is also defined as conditional on \( y \). In this setting, the ELBO is adapted as shown in Equation~\ref{eq:elbo_cond}. As in the standard VAE case, this expression comprises two main terms: the reconstruction error, which measures the model’s ability to recover the input \( x \) from the latent representation \( z \) and the condition \( y \); and the KLD term, which penalizes deviations between the approximate posterior \( q_{\phi}(z|x,y) \) and the conditional prior \( p(z|y) \). Minimizing the KLD enables the CVAE to learn a structured latent space that is coherent with the conditioning variable, thereby making it possible to generate realistic samples by drawing \( z \sim p(z|y) \) and decoding it together with \( y \).

\begin{equation}
\begin{split}
\log p(x|y) \geq\; & \mathbb{E}_{q_{\phi}(z|x,y)}[\log p_{\theta}(x|z,y)] \\
& - \beta \, \text{KLD}(q_{\phi}(z|x,y) \parallel p(z|y))
\end{split}
\label{eq:elbo_cond}
\end{equation}

\section{Cardiac rehabilitation process}
To carry out the data generation and cardiac risk prediction tasks addressed in this work, a dataset constructed from a cardiac rehabilitation program offered by the Cardiology Department of the CHUS (University Hospital Complex of Santiago de Compostela, Spain) was used. The dataset comprises 370 variables that monitor the status of 3,211 anonymized patients during and after completion of the program. All individuals included in the database had previously experienced some type of cardiovascular event or presented a high risk of heart disease. However, not all of them ultimately agreed to participate in the program, and therefore complete data are not available for all patients.

The variables collected in the study can be grouped into several clinical and functional categories. Some are common to any analysis of patients with cardiovascular disease, while others are specific to cardiac rehabilitation. The main categories are:

\begin{itemize}
    \item \textbf{Sociodemographic and anthropometric data:} basic information such as age, employment status, sex, weight, height, waist circumference, among others.

    \item \textbf{Previous clinical history:} presence of prior medical conditions, such as arterial hypertension, atrial fibrillation, diabetes mellitus, anemia, or other relevant chronic diseases.

    \item \textbf{Biomarkers and diagnostic tests:} cholesterol, glucose, creatinine, triglycerides, potassium, blood pressure, etc.

    \item \textbf{Functional capacity and response to exercise:} data obtained from exercise stress tests (ergometry) to assess cardiac response to progressive physical exertion. This group includes four variables whose evolution determines whether the patient remains at risk at the end of the program:
    \begin{itemize}
        \item \textbf{VO\textsubscript{2} peak:} measures the maximum oxygen consumption during exercise, reflecting cardiorespiratory capacity.
        \item \textbf{Watts:} indicates the physical power generated during the stress test, related to muscular performance.
        \item \textbf{METs:} represent exercise intensity relative to resting energy expenditure.
        \item \textbf{Exercise duration (minutes):} reflects the total duration of the test until the exercise tolerance limit is reached.
    \end{itemize}
        
    \item \textbf{Clinical events during follow-up:} episodes such as acute myocardial infarction, arrhythmias, hypoglycemic events, emergency department visits, hospital admissions, program start and end dates, follow-up visits, etc.

    \item \textbf{Mental health and lifestyle questionnaires:} results from standardized questionnaires assessing depressive symptoms, anxiety, smoking habits, physical activity, among others.

    \item \textbf{Pharmacological treatment:} records of prescribed medication, including anticoagulants, antiplatelet agents, statins, insulin, etc.
\end{itemize}

\subsection{Main limitations}
Despite the richness and variety of the available variables, the dataset presents several limitations that led to changes in the initially conceived approach and that must be considered when interpreting the results presented in this thesis. First, out of the initial 3,211 patients, only 811 meet the minimum requirements for the intended analysis, which consist of having two measurements (baseline and final) for at least one of the four variables used for cardiac risk prediction (VO$_2$ peak, Watts, METs, and exercise duration).

Furthermore, during the analysis process, inconsistencies, impossible values, transcription errors, and a large number of missing data were detected, attributable to clinical variability among patients. As a result, the amount of valid data available is considerably lower than expected. In addition, the fact that different events are recorded using a single date variable per event (overwritten if another event occurs) leads to a substantial loss of valuable information for analysis. In practice, experiencing multiple myocardial infarctions is not equivalent to experiencing only one; however, in the dataset both situations are treated identically. Similarly, if the last event of a given type occurred after the end of the program, it is not possible to determine whether the patient experienced such an event during the period of interest.

Finally, the main identified issue is that most events occurring throughout the program, such as cardiovascular events, episodes involving physiological alterations (hypertension, hypoglycemia, etc.), or emergency admissions, do not have any associated measurements or variables beyond the date on which they occurred. This severely degraded the results that could be obtained by treating the rehabilitation program as a business process, since these events present null values in the trace attributes. Variables generally contained values only for four events: prior admission (only for a subset of attributes), program start, program end, and the follow-up visit six months after program completion. As the objective is to predict cardiac risk immediately before program completion, only measurements from the initial admission and program start can be used, without the possibility of observing session-by-session evolution.

\section{Methodology}
\subsection{Data Preparation}
The approach proposed in this study targets the improvement of cardiac risk prediction at the end of a rehabilitation program using a dataset characterized by a limited number of valid instances, substantial missing data due to heterogeneous data collection, and the presence of weakly informative variables. Consequently, a rigorous data preparation pipeline is required to ensure data quality and analytical suitability. This process integrates domain-specific medical knowledge and comprises the following key stages:

\begin{itemize}
\item \textbf{Selection of relevant records:} Patients with sufficient information for analysis were selected, specifically those who completed the rehabilitation program and presented both initial and final measurements. Since cardiac risk is defined based on changes in VO$_2$ peak, METs, Watts, and exercise duration, at least one of these variables had to be available at both time points.

\item \textbf{Preliminary variable selection:} Variables collected after program completion were excluded to avoid information leakage. Among end-of-program measurements, only the four exercise test variables required for defining cardiac risk were retained for generation purposes and excluded from the prediction stage, while the remaining end variables were discarded.

\item \textbf{Data cleaning:} Selected variables were reviewed to identify and correct evident inconsistencies or out-of-range values when possible; otherwise, unreliable entries were removed to prevent the introduction of unrealistic information.

\item \textbf{Trace-based representation:} Treating cardiac rehabilitation as a business process, each patient’s clinical evolution was encoded as a single process trace, represented in tabular form. Each row corresponds to a patient, capturing mandatory events (admission, program start, and completion) and relevant intermediate clinical events. This representation was chosen to avoid sparsity issues arising from unavailable measurements associated with intermediate events.

\item \textbf{Discretization of continuous variables:} Continuous variables were converted into categorical intervals, with an additional category reserved for missing values. This strategy avoids medical imputation, preserves clinically meaningful absence patterns, reduces sensitivity to outliers, and mitigates overfitting~\cite{DBLP:journals/access/ZaidiDW20}.

\item \textbf{Feature selection and engineering:} Clinically informed feature selection was applied to remove variables with excessive missingness or high redundancy. Correlation analysis using the Pearson coefficient~\cite{pearson_1895} eliminated highly correlated variables (absolute correlation $>0.9$)~\cite{dormann2013collinearity}. Additional variables were derived through feature engineering, including BMI computation, temporal variables expressed as weeks since program start, and the definition of the cardiac risk label based on a 15\% improvement threshold.
\end{itemize}

These steps yield a refined dataset whose characteristics are summarized in Table~\ref{tab:resumo_estrutura}. Two final preprocessing stages complete the preparation for generative modeling. First, categorical variables are encoded numerically to ensure compatibility with neural networks. Second, the data are partitioned into mini-batches, which are sequentially processed during CVAE training. Mini-batch learning, widely adopted in state-of-the-art approaches~\cite{DBLP:conf/nips/XuSCV19, DBLP:conf/icml/KotelnikovBRB23, DBLP:journals/ai/WangN25}, enables efficient computation, improves generalization, and promotes stable convergence.

\begin{table*}[ht]
\centering
\begin{tabular}{ccccc}
\toprule
\textbf{No. of instances} & \textbf{No. of variables} & \textbf{Categorized} & \textbf{Categorical} & \textbf{Binary} \\
\midrule
811 & 91 & 52 & 16 & 23 \\
\bottomrule
\end{tabular}
\caption{Structural summary of the dataset after preprocessing.}
\label{tab:resumo_estrutura}
\end{table*}

\subsection{Proposed Architecture}
Once the data has been properly prepared, the next step is to design the architecture to be used for synthetic data generation. As previously indicated, this architecture is based on a CVAE. The choice of this architecture is motivated by its ability to support conditional generation, allowing explicit control over the synthesis process and the generation of patient traces with predefined characteristics. Additionally, architectures based on VAE and CVAE exhibit more stable training compared to GANs, and they allow a more direct handling of categorical variables, which are used in this work.

The proposed architecture is a CVAE adapted for tabular data, as each trace is represented as a row due to the characteristics of the dataset. The CVAE, through the condition, incorporates additional information into both the encoding and reconstruction processes, influencing the encoder and the decoder. Specifically, the condition represents the class associated with the patient, that is, whether the patient is at risk or has improved during the program.

\begin{figure*}[ht]
  \centering
  \subfloat[Encoder]{%
    \includegraphics[width=0.51\linewidth]{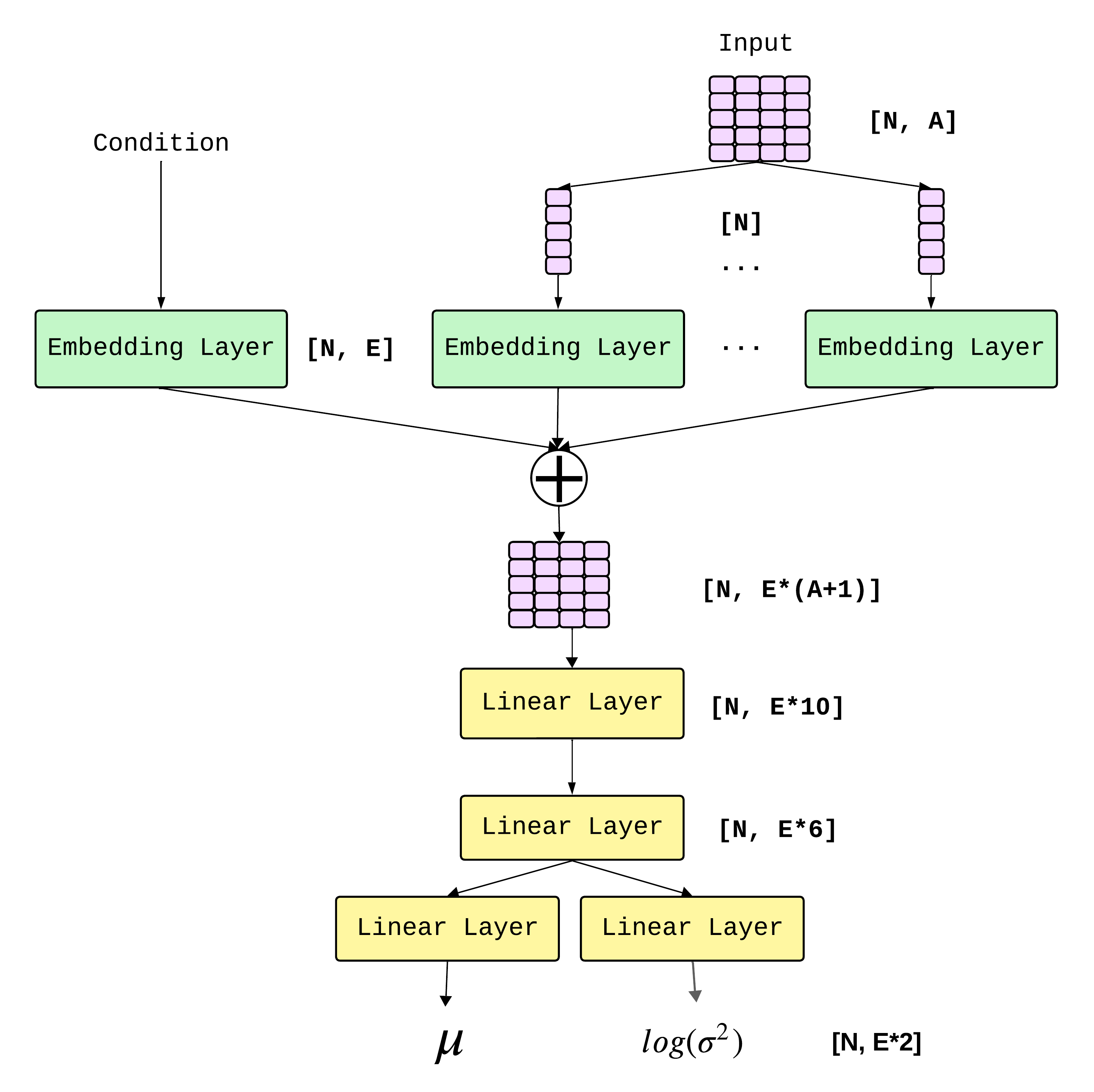}%
    \label{fig:arqenc}%
  }
  \hfill
  \subfloat[Decoder]{%
    \includegraphics[width=0.46\linewidth]{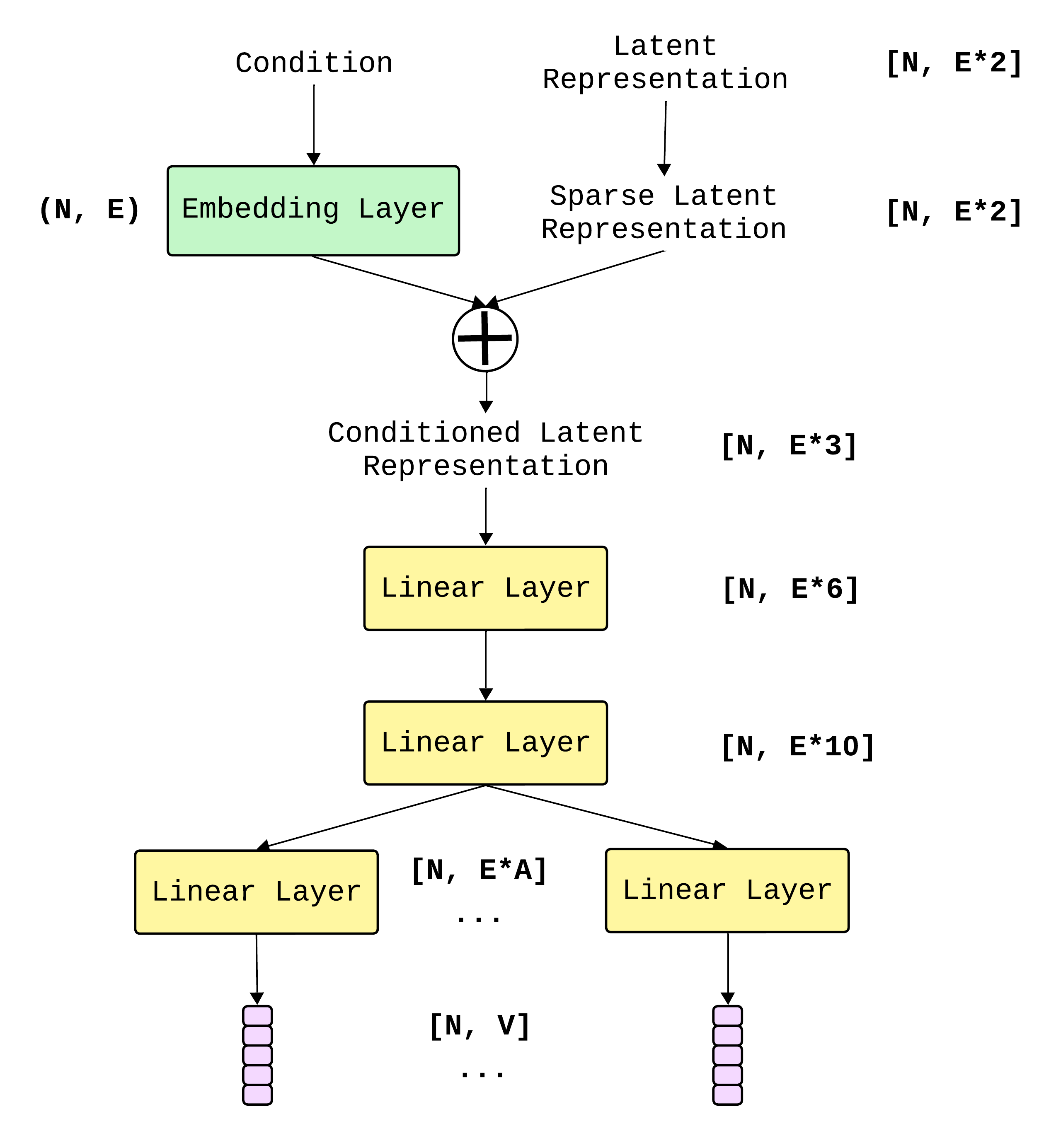}%
    \label{fig:arqdec}%
  }
  \caption{Architecture of the proposed CVAE.}
  \label{fig:arqF}
\end{figure*}

Figure~\ref{fig:arqF} shows the final proposed architecture, which will be described in detail in the following subsections. Black annotations indicate the dimensions of the data at each stage, where \(N\) is the mini-batch size, \(A\) is the number of attributes, \(E\) is the embedding dimension, and \(V\) is the number of distinct values for each attribute. As observed, the layer sizes depend on other parameters such as the number of attributes or embedding dimension, facilitating model adaptation to different input configurations and reducing the number of hyperparameters to tune.

It is worth noting that this was the architecture that yielded the best results, although it was not the only one tested. Section~\ref{sec:metodo-experimental} presents other designed variants, highlighting the differences from the selected main architecture.

\subsubsection{Encoder}
The encoder, depicted in Fig.~\ref{fig:arqenc}, transforms the input data into a compressed latent representation that captures the most relevant features of patient traces. It receives as input a mini-batch of patient data together with a batch of conditional information indicating patient risk evolution. Each input attribute is processed independently and mapped to a dedicated embedding space, enabling an efficient numerical representation of categorical variables while preserving semantic relationships among categories.

The resulting attribute embeddings are concatenated and combined with an embedded representation of the condition, forming a unified input that is subsequently passed through a series of linear transformations with non-linear activation functions. These layers progressively reduce the dimensionality of the representation, yielding a latent encoding that summarizes the essential characteristics of the input data.

From this encoding, two separate linear layers compute the parameters of the latent probability distribution: the mean \( \mu \) and the log-variance \( log(\sigma^2)\). These parameters define the conditional latent distribution from which a sample is drawn using the reparameterization trick, ensuring stochasticity while preserving differentiability during training:
\begin{equation}
\mathbf{z} = \mu + \exp\!\left(\tfrac{1}{2} log(\sigma^2)\right)\cdot \epsilon,
\quad \epsilon \sim \mathcal{N}(0,1).
\end{equation}

\subsubsection{Decoder}
The decoder (Fig.~\ref{fig:arqdec}) reconstructs the input data from the latent representation and is designed symmetrically to the encoder. To promote compact and interpretable latent representations, an \(L_1\) regularization term is applied to the sampled latent vectors~\cite{DBLP:journals/corr/abs-2405-14270}, encouraging the model to focus on the most informative latent dimensions, reducing noise and improving generalization. The sparsified latent vector is then concatenated with the embedded condition to form a conditioned latent representation, which the decoder uses to generate more realistic and diverse synthetic samples.


This representation is then passed through a sequence of linear layers with non-linear activations that progressively expand the feature space. The final decoder stage consists of multiple parallel output layers, one per input attribute, each producing a categorical distribution over the possible values of that attribute. This design enables attribute-wise reconstruction while preserving flexibility and scalability with respect to the number and nature of input variables.

During training, the decoder outputs the reconstructed attributes jointly with the latent distribution parameters provided by the encoder, constituting the complete output of the CVAE model.

\subsection{Synthetic Sample Generation via Sampling}
In a generative model such as a CVAE, sample generation consists of drawing latent vectors \( \mathbf{z} \) from the learned latent space to produce new data via the decoder. These latent vectors are obtained through a sampling process, typically from a standard normal distribution or using the parameters (mean and variance) learned by the encoder. This allows controlled exploration of the latent space to generate new samples similar to those in the training set.

In this work, a latent-space-guided sampling strategy is employed. Instead of merely drawing random samples from the Gaussian distribution learned by the model, this approach is combined with the SMOTE technique (Synthetic Minority Oversampling Technique)~\cite{DBLP:journals/jair/ChawlaBHK02, DBLP:conf/ijcnn/HeBGL08}, a widely used interpolation method for synthetic data generation. Specifically, new latent vectors are generated by interpolating between real representations extracted by the encoder, using Euclidean distance to select the nearest neighbors. This strategy allows the generation of more realistic and representative synthetic data by preserving local coherence in the latent space. It also promotes better coverage of regions associated with underrepresented data, incorporating structural information learned during training.

\subsection{Training}
The training loss is composed of two components: one due to the CVAE and another associated with contrastive learning.

\vspace{2mm}
\noindent \textbf{CVAE Loss.} CVAEs are trained with a dual objective: on one hand, to accurately reconstruct the input data, as a traditional autoencoder would, and on the other, to approximate the distribution learned by the encoder to a prior distribution, usually a standard normal. Therefore, this loss, corresponding to the ELBO, consists of a reconstruction term and a term associated with the divergence between distributions, as shown in Equation~\ref{eq:elbo2}.

\begin{equation}
\mathcal{L}_{CVAE} = \text{CE}(y, \hat{y}) - \text{KLD}(q_{\phi}(z|x,y) \parallel p(z|y))
\label{eq:elbo2}
\end{equation}

Since all variables used are categorical, the reconstruction term is calculated using cross-entropy, which measures the discrepancy between the predicted probability distributions (\( \hat{y}_i \)) and the true label distributions (\( y_i \)), according to Equation~\ref{eq:loss_ce}. Here, \(C\) represents the number of possible classes for a given attribute.

\begin{equation}
    \text{CE}(y, \hat{y}) = - \sum_{i=1}^{C} y_i \log(\hat{y}_i) \label{eq:loss_ce}
\end{equation}

Complementarily, the Kullback-Leibler divergence~\cite{DBLP:books/daglib/0016881} measures the difference between the learned latent distribution and the standard Gaussian. For two arbitrary normal distributions, the formula is expressed following Equation~\ref{eq:loss3}:

\begin{equation}
    \mathrm{KLD}[p, q] = \frac{1}{2} \left[ \frac{(\mu_p - \mu_q)^2}{\sigma_q^2} + \frac{\sigma_p^2}{\sigma_q^2} - \ln\left(\frac{\sigma_p^2}{\sigma_q^2}\right) - 1 \right] \label{eq:loss3}
\end{equation}

When \( q = \mathcal{N}(0,1) \), this simplifies to:

\begin{equation}
    \text{KLD}[p, \mathcal{N}(0,1)] = \frac{1}{2} \left( 1 + \log(\sigma^2) - \mu^2 - \sigma^2 \right) \label{eq:loss4}
\end{equation}

\vspace{2mm}
\noindent \textbf{Contrastive Loss.} To promote a more structured and informative latent representation, a contrastive loss term is added based on contrastive learning~\cite{DBLP:conf/cvpr/HadsellCL06}. This unsupervised technique encourages similar samples to remain close in latent space while dissimilar ones are pushed apart, which is especially useful in generative models like the CVAE to achieve a well-structured and semantically coherent latent space. The employed contrastive loss is the InfoNCE loss~\cite{DBLP:journals/corr/abs-1807-03748}, defined in Equation~\ref{eq:contrastive}:

\begin{equation}
\mathcal{L}_{\text{NCE}} = -\log \frac{\exp(\mathrm{sim}(\mathbf{z}, \mathbf{z}^{+}) / \tau)}{\sum\limits_{j=1}^{K} \exp(\mathrm{sim}(\mathbf{z}, \mathbf{z}_{j}) / \tau)} \label{eq:contrastive}
\end{equation}

Here, \( \mathbf{z} \) and \( \mathbf{z}^{+} \) represent a positive pair (e.g., two views of the same sample), while \( \mathbf{z}_{j} \) are negative examples (different samples). The function \( \mathrm{sim}(\cdot, \cdot) \) measures similarity between vectors, and \( \tau \) is a temperature parameter controlling dispersion. This formulation encourages similar latent representations to remain close while separating distinct ones, improving the semantic organization of the latent space~\cite{DBLP:conf/cikm/0158ZLYY22}.

\vspace{2mm}
\noindent \textbf{Total Loss.} The total loss function is a weighted combination of the ELBO and the contrastive loss, as shown in Equation~\ref{eq:total}:

\begin{equation}
\mathcal{L}_{\text{total}} = \mathcal{L}_{\text{CE}} + \beta \cdot \mathcal{L}_{\text{KLD}} + \alpha \cdot \mathcal{L}_{\text{NCE}}
\label{eq:total}
\end{equation}

Here, \( \beta \) and \( \alpha \) are hyperparameters weighting the importance of regularization and contrastive learning, respectively. Increasing \(\alpha\) emphasizes the contrastive loss, promoting more discriminative and structured latent representations. However, excessively high values can hinder the VAE's ability to reconstruct data accurately, creating a trade-off between fidelity and latent space separation.

A cyclical adjustment is proposed for \(\beta\), which regulates the weight of the Kullback-Leibler divergence, to prevent premature latent space collapse at the beginning of training~\cite{DBLP:conf/naacl/FuLLGCC19}. This phenomenon occurs because the decoder initially receives uninformative latent codes, causing the KLD term to quickly drop to zero, and the model ignores the latent vector, losing the ability to learn useful representations. With this strategy, \(\beta\) gradually increases from 0 to 1 over several epochs, repeating the process for a defined number of cycles. This alternates phases where the model freely explores the latent space (\(\beta\) low) and phases with stronger regularization (\(\beta\) high), promoting better organization of the latent space while preserving the model's capacity to capture complex patterns in the input data.

During each training iteration, the total loss is computed, and model weights are updated via backpropagation to minimize it. For stable and efficient training, patient traces are processed in fixed-size mini-batches, with weight updates performed after each batch. Training continues for \(e\) epochs or early stopping is applied if no improvement in loss is observed for 10 consecutive epochs.
\subsection{Experimental Design}
After designing and implementing the architecture, it is necessary to verify whether it effectively generates coherent data, increasing both the richness and size of the training dataset, with the ultimate goal of improving the prediction of patients’ cardiac risk at the end of the program.

The validation of the proposed generative model is based on an indirect approach, consisting of evaluating its impact on the performance of classifiers responsible for determining whether a patient is at risk or not at the end of the program. Rather than directly measuring the quality of the generated sequences, the analysis focuses on whether they contribute to improving the predictive capability of other models when included as additional training data.

Initially, a set of classifiers is trained using a training set composed exclusively of real data to estimate cardiac risk. The original dataset is split into 80\% for training and 20\% for final testing. Additionally, the training subset is further split into 80\% for actual training and 20\% for validation, allowing hyperparameter tuning and preventing overfitting. The selected classifiers are Random Forest, XGBoost, TabNet~\cite{DBLP:conf/aaai/ArikP21}, and TabTransformer~\cite{DBLP:journals/corr/abs-2012-06678}, chosen for their effectiveness with tabular data.

Once the baseline performance of these models is evaluated on the validation and test sets, a data augmentation phase is introduced. The training set is expanded by adding synthetic sequences generated by the CVAE. The classifiers are retrained on this enriched dataset, keeping the validation and test sets unchanged, to assess whether the generated information improves the models’ ability to learn relevant patterns for cardiac risk prediction.

\subsubsection{Metrics}
Identifying patients at risk versus not at risk constitutes an imbalanced classification problem, as most patients exit the risk condition by the end of the program. In this context, metrics like accuracy may be misleading, as correctly classifying the majority class while misclassifying the minority class can still yield a high score. The real challenge lies in accurately identifying both patients who do not improve during the program and those who remain at risk. Consequently, a more appropriate metric is the F-score, which combines precision and recall. Precision measures the proportion of true positives among all instances predicted as positive. Recall measures the proportion of true positives among all actual positive instances in the dataset. The F1-score, used in this study, gives equal weight to both metrics and is defined in Equation~\ref{eq:fs1}:

\begin{equation}
\text{F}_{1}\!\!-\!\text{score} = 2 \cdot \frac{\text{Precision} \cdot \text{Recall}}{\text{Precision} + \text{Recall}} \label{eq:fs1}
\end{equation}

The same procedure is followed to compare the proposed architecture with state-of-the-art generative models~\cite{DBLP:conf/nips/XuSCV19, DBLP:journals/ai/WangN25, DBLP:conf/nips/GulrajaniAADC17}, evaluating the effectiveness of each in generating synthetic data. In particular, the impact of the generated augmentation on classifier performance is compared to determine which approach provides the most beneficial improvement for cardiac risk prediction.

Another crucial aspect of conditional generative models is their ability to generate samples that are not only realistic but also consistent with the class label provided as a condition. This is particularly relevant in sensitive domains such as healthcare. The clinical conditions distinguishing patients at risk and not at risk are well defined: the latter show at least a 15\% improvement in one of the four ergometry variables (VO$_2$ peak, Watts, METs, and exercise duration), whereas a patient at risk shows no significant improvement. Therefore, it is possible to verify whether synthetic instances generated for a given class \( \hat{x} \in C_i \) meet the real criteria. For this task, the class-specific accuracy can be computed for models that allow conditional generation:

\[
\text{Accuracy}_{C_i} = \frac{\# \{ \hat{x} \in C_i : \hat{x} \text{ meets the conditions} \}}{\# \{ \hat{x} \in C_i \}}
\]

\subsubsection{Defined Models}\label{sec:metodo-experimental}
The main proposed model, named \textbf{Sparse Contrastive CVAE} (SCCVAE), had its key parameters manually selected to balance realistic and consistent generation with the characteristics of the cardiac rehabilitation dataset. First, embedding dimensions were set to \( s_{\text{emb}} = 32 \), which in turn determines the layer sizes in both the encoder and decoder. This embedding size balances representation capacity and efficiency, capturing relevant information without excessively costly training. The latent space dimension was fixed at \( h = 64 \), as lower values led to poor representations, negatively impacting the quality of generated data, while higher values increased computational cost without significant improvements. For the cyclical adjustment of \(\beta\), four cycles (\(cycles=4\)) with a ratio of 0.9 were used, meaning \(\beta\) increases during 90\% of the cycle and remains at its maximum (\(\beta_{\max} = 1\)) only for the remaining 10\%, prioritizing reconstruction.

The temperature parameter in the contrastive loss is key to controlling the differentiation between similar and distinct latent representations, directly affecting the model's ability to effectively cluster related samples. A temperature of \( \tau = 0.5 \) was chosen, as higher values reduced class discrimination, while lower values made training unstable. The weighting parameter \( \alpha \) for the contrastive loss was set to \( \alpha = 0.1 \), balancing this regularization term with the ELBO components, avoiding dominance of the contrastive loss over the main model objectives. Additionally, an \( L_1 \) penalty was applied to model weights with \( \lambda = 10^{-3} \) to encourage sparsity, reducing overfitting and improving interpretability while maintaining generalization.

Training was carried out for up to \( e = 200 \) epochs, with early stopping applied if the loss did not improve for 10 consecutive epochs.

Several variants of this architecture were also developed to analyze the impact of specific components or techniques on final performance:
\begin{itemize}
    \item \textbf{SCVAE}: This variant removes the contrastive loss, under the hypothesis that it may induce overly artificial clusters, leading to erroneous or distorted representations.
    \item \textbf{CCVAE}: This variant eliminates the \( L_1 \) penalty, allowing greater freedom in the weights to assess if this improves the model's ability to capture complex relationships, at the cost of increased overfitting risk.
    \item \textbf{SCVAE-C$\alpha$}: In this variant, the \(\alpha\) parameter varies cyclically between 0 and 1 during training, following a pattern similar to \(\beta\) with \(cycles=4\) and \(ratio=0.9\), modulating the contrastive regularization weight to enhance latent structure while avoiding domination of the loss function.
\end{itemize}

These defined architectures are also compared with three state-of-the-art generative models with publicly available implementations: CTGAN~\cite{DBLP:conf/nips/XuSCV19}, TTVAE~\cite{DBLP:journals/ai/WangN25}, and WGAN-GP~\cite{DBLP:conf/nips/GulrajaniAADC17}. Hyperparameters for each followed the settings provided in the original publications.

Finally, for each classifier considered, an exhaustive grid search over a predefined set of hyperparameters was performed, selecting the configuration that yielded the best validation performance. This ensures a fair and optimized evaluation of each model, both on the original and augmented datasets. 

\section{Results and Discussion}
Table~\ref{tab:f1_scores_completa} presents the F1-score results obtained by the different classifiers used to predict patients’ cardiac risk at the end of the rehabilitation program. Specifically, it includes both the F1-score for the at-risk class, which is of particular interest as the minority class, and the weighted F1-score (\( F_1^{\text{w}} \)), representing the class-size-weighted average. For each model, results are shown for the original training set (533 samples) as well as for datasets augmented using the architectures proposed in Section~\ref{sec:metodo-experimental}. In each case, the training set was augmented by factors of 2 and 5 over the base size (1066 and 2665 samples, respectively), maintaining the initial class distribution. For each classifier, the three datasets yielding the best performance are highlighted in three shades of green, with the darkest shade indicating the highest score.
\begin{table*}[tb]
\centering
\small
\setlength\dashlinedash{2pt}
\setlength\dashlinegap{1.5pt}
\setlength\arrayrulewidth{0.3pt}
\begin{tabular}{lcccccccc}
\toprule
& \multicolumn{2}{c}{\textbf{XGBoost}} 
& \multicolumn{2}{c}{\textbf{Random Forest}} 
& \multicolumn{2}{c}{\textbf{TabNet}} 
& \multicolumn{2}{c}{\textbf{TabTransformer}} \\
\cmidrule(lr){2-3} \cmidrule(lr){4-5} \cmidrule(lr){6-7} \cmidrule(lr){8-9}
& \( \mathbf{F}_1 \) risk & \( F_1^{\text{w}} \) 
& \( \mathbf{F}_1 \) risk & \( F_1^{\text{w}} \)
& \( \mathbf{F}_1 \) risk & \( F_1^{\text{w}} \) 
& \( \mathbf{F}_1 \) risk & \( F_1^{\text{w}} \) \\
\midrule
\textbf{Original data}             & 0.6418 & 0.7133 & 0.6107 & 0.6942 & 0.5401 & 0.6249 & 0.5484 & 0.6602 \\
\cdashline{1-9}
\textbf{SCCVAE *2}                   & \cellcolor{green!80!black}0.7153 & \cellcolor{green!80!black}0.7678 & \cellcolor{green!80!black}0.6767 & \cellcolor{green!80!black}0.7428 & \cellcolor{green!80!black}0.5672 & 0.6535 & \cellcolor{green!20!white}0.5667 & \cellcolor{green!20!white}0.6818 \\
\textbf{SCCVAE *5}                   & 0.6923 & \cellcolor{green!70!white}0.7598 & \cellcolor{green!70!white}0.6615 & \cellcolor{green!70!white}0.7358 & \cellcolor{green!20!white}0.5574 & \cellcolor{green!70!white}0.6710 & \cellcolor{green!80!black}0.5833 & \cellcolor{green!80!black}0.6940 \\
\cdashline{1-9}
\textbf{SCVAE *2}                    & 0.6557 & 0.7441 & 0.6102 & 0.7171 & 0.5441 & 0.6305 & 0.5345 & 0.6662 \\
\textbf{SCVAE *5}                    & 0.6457 & 0.7285 & \cellcolor{green!20!white}0.6475 & 0.7087 & 0.5528 & 0.6642 & 0.5649 & 0.6582 \\
\cdashline{1-9}
\textbf{CCVAE *2}                    & 0.6562 & 0.7350 & 0.6190 & 0.7099 & 0.5440 & 0.6548 & 0.5172 & 0.6539 \\
\textbf{CCVAE *5}                    & 0.6617 & 0.7309 & 0.6412 & 0.7182 & \cellcolor{green!70!white}0.5625 & 0.6627 & 0.5620 & 0.6764 \\
\cdashline{1-9}
\textbf{SCCVAE -\(\mathbf{C\alpha}\) *2}      & \cellcolor{green!20!white}0.6957 & 0.7501 & 0.6271 & \cellcolor{green!20!white}0.7294 & 0.5528 & \cellcolor{green!20!white}0.6656 & 0.5517 & 0.6786 \\
\textbf{SCCVAE -\(\mathbf{C\alpha}\) *5}          & \cellcolor{green!70!white}0.6963 & \cellcolor{green!20!white}0.7554 & 0.6230 & 0.7198 & 0.5546 & \cellcolor{green!80!black}0.6749 & \cellcolor{green!70!white}0.5738 & \cellcolor{green!70!white}0.6832 \\
\cdashline{1-9}
\textbf{CTGAN *2}                   & 0.6565 & 0.7302 & 0.6269 & 0.7013 & 0.4404 & 0.6149 & 0.3738 & 0.5740 \\
\textbf{CTGAN *5}                    & 0.6187 & 0.6849 & 0.5806 & 0.6845 & 0.3697 & 0.5399 & 0.3019 & 0.5278 \\
\cdashline{1-9}
\textbf{TTVAE *2}                    & 0.6179 & 0.7143 & 0.5714 & 0.6872 & 0.4538 & 0.6013 & 0.3860 & 0.5649 \\
\textbf{TTVAE *5}                    & 0.5484 & 0.6602 & 0.5500 & 0.6695 & 0.3333 & 0.5276 & 0.2857 & 0.4998 \\
\cdashline{1-9}
\textbf{WGAN-GP *2}                      & 0.6562 & 0.7350 & 0.6364 & 0.7126 & 0.4961 & 0.6091 & 0.4954 & 0.6527 \\
\textbf{WGAN-GP *5}                      & 0.5586 & 0.6926 & 0.6395 & 0.6861 & 0.3922 & 0.5981 & 0.4000 & 0.6078 \\
\bottomrule
\end{tabular}
\caption{\( F_1 \) score for the at-risk class and weighted \( F_1 \) for different generators and classifiers.}
\label{tab:f1_scores_completa}
\end{table*}

As observed in Table~\ref{tab:f1_scores_completa}, \textbf{SCCVAE} consistently achieves the best results compared to the other evaluated generators, closely followed by SCCVAE -\(\mathbf{C\alpha}\). Regardless of the classifier used, synthetic data generated by SCCVAE achieves higher scores than those obtained with the original data, demonstrating its ability to capture and reinforce the relevant structures of the dataset. Quantitatively, doubling the training set size with SCCVAE yields improvements greater than 0.04 in both the \( F_1 \) for the at-risk class and the \( F_1^{\text{w}} \) for Random Forest and XGBoost classifiers, reaching for the latter an \( F_1 \) risk score of 0.7153 and an \( F_1^{\text{w}} \) of 0.7678, the overall best results, corresponding to improvements of 0.07 and 0.05, respectively. For TabNet and TabTransformer, the larger augmented set achieves the best results, with a 0.03 improvement over the base data. Data generated by SCCVAE consistently ranks among the top three options across almost all classifiers, highlighting its potential as a robust and effective tool for generating high-quality synthetic data in this problem domain.

Regarding the other generative models, SCVAE and CCVAE variants show intermediate performance: they improve results compared to the original data (except for the CCVAE double-size set for TabTransformer) but with less consistency and impact than SCCVAE. This indicates that both sparsity and contrastive learning positively contribute to the CVAE data generation process, and their combination yields the two best-performing generative models. In contrast, more generic generators such as CTGAN, TTVAE, and WGAN-GP exhibit noticeably lower performance than the VAE-based variants proposed here. In many cases, they fail to surpass the original data results and never rank among the best-performing sets, suggesting a limited ability to capture the relevant characteristics of this dataset.

It is worth noting that, in many instances, classifiers trained on doubled-size synthetic sets outperform those trained on quintupled sets. This is because excessive generation may introduce noise into the dataset. Creating large amounts of artificial samples increases the likelihood of including unrealistic or ambiguous examples, which can hinder learning and reduce model generalization. Therefore, generating excessive synthetic samples does not always lead to improved performance.

However, we are not only interested in improving the performance of the final classifier, but also in the generative architecture’s ability to produce coherent and realistic samples. Since the characteristics defining at-risk and non-risk patients are known, it is possible to evaluate whether the generative model can create instances that truly satisfy these specific conditions. Specifically, it is verified that at-risk participants do not show an increase greater than 15\% in any of the ergometry variables, while non-risk participants indeed show improvements above this threshold. This step is crucial to ensure that the model generates not only superficially plausible data but also respects the semantics of each class, guaranteeing the quality and utility of the generated samples for downstream tasks. Table~\ref{tab:acerto_modelos} presents the accuracy of the different generative models for each class.

\begin{table*}[htbp]
\centering
\begin{tabular}{lccccccc}
\toprule
 & SCCVAE & SCVAE & CCVAE & SCCVAE-$C\alpha$ & CTGAN & TTVAE & WGAN-GP \\
\midrule
$\mathrm{Accuracy}_{C_{\mathrm{risk}}}$     & 100.00\% & 100.00\% & 98.84\% & 100.00\% & 24.53\% & - & - \\
$\mathrm{Accuracy}_{C_{\mathrm{non-risk}}}$ & 100.00\% & 99.38\% & 100.00\% & 100.00\% & 78.82\% & - & - \\
\bottomrule
\end{tabular}
\caption{Accuracy results for at-risk and non-risk cases according to different generative models.}
\label{tab:acerto_modelos}
\end{table*}

As observed, the proposed CVAE-based models in this work are capable of generating synthetic examples that respect the desired class (risk or non-risk), reinforcing the strong results shown in Table~\ref{tab:f1_scores_completa}, as they maintain the original data distribution. In contrast, more generic models, CTGAN, TTVAE, and WGAN-GP, show significant limitations: the latter two do not support conditional generation, making it difficult to control the generated class, while CTGAN, although conditional, achieves low accuracy for the at-risk class, altering the original distribution. This mismatch between the intended condition and the actual generated samples suggests that these models introduce a substantial amount of noise into the synthetic data, blurring class-specific patterns. These factors explain their poor classification performance and highlight the importance of preserving the conditions of each class.

\section{Conclusion}
In this work, a novel deep learning-based generative architecture was proposed for predicting cardiac risk. The motivation behind this proposal stems from modeling the cardiac rehabilitation program as a business process, aiming to capture patients’ evolution over time. Specifically, a Conditional Variational Autoencoder (CVAE) was designed, adapted to encode the clinical features collected at different phases of the program. Additionally, data generation is conditioned on the risk class, allowing the model to capture not only the internal structure of the information but also to generate synthetic samples that are more realistic and clinically coherent.

This architecture generates new synthetic physiological traces while maintaining consistency with real data distributions. Regularization mechanisms, such as contrastive loss and \(L_1\) penalty, are proposed to improve latent representations and promote higher diversity and realism in the traces. Moreover, a guided sampling methodology in the latent space via interpolation is developed to create more representative samples in low-density regions. This strategy expands latent space coverage and enhances sample diversity, especially in the context of class imbalance.

The results demonstrate that including data generated by the proposed architecture improves cardiac risk prediction across multiple models, outperforming state-of-the-art alternatives by producing more useful and representative synthetic data. Nevertheless, there is room for improvement. Future work could explore the use of reference distributions more complex than $ \mathcal{N}(0, 1) $, deeper architectures or hierarchical latent spaces, and new regularization methods to enhance the model’s generative capacity. Additionally, having a more complete and information-rich dataset that includes variables associated with all relevant events in the rehabilitation program would be beneficial. This would allow patient traces to form more comprehensive sequential structures, capturing the clinical process evolution more broadly and in greater detail.


%



\section*{Acknowledgment}
This work has received financial support from the Consellería de Educación, Universidade e Formación Profesional (accreditation 2024-2027 ED431G-2023/04), the European Regional Development Fund (ERDF), which acknowledges the CiTIUS - Centro Singular de Investigación en Tecnoloxías Intelixentes da Universidade de Santiago de Compostela as a Research Center of the Galician University System, and the Spanish Ministry of Science and Innovation (grants PID2020-112623GB-I00, TED2021-130374B-C21 and PID2023-149549NB-I00).





\bibliographystyle{IEEEtran}
\bibliography{biblio}
\end{document}